\documentclass[10pt,twocolumn,letterpaper]{article}

\usepackage{wacv}
\usepackage{times}
\usepackage{epsfig}
\usepackage{graphicx}
\usepackage{amsmath}
\usepackage{amssymb}
\usepackage{hyperref}

\wacvfinalcopy % *** Uncomment this line for the final submission

\def\wacvPaperID{****} % *** Enter the wacv Paper ID here

\ifwacvfinal\fi
\begin{document}

%%%%%%%%% TITLE
\title{Sequence Information Channel Concatenation for Improving Camera Trap Image Burst Classification}
%TODO: Can we have a better title? 
% - Exploring Short Burst of Sequences for Improving Camera Trap Image Classification
% - Sequence Information Channel Concatenation for Improving Camera Trap Image Classification
% - Sequence Information Channel Concatenation for Improving Camera Trap Image Burst Classification
%
% DSM: IMO "Sequence Information Channel Concatenation" is confusing, and you also say both "Burst" and
% "Sequence" in the current title, which mean the same thing.
%
% Some alternatives:
%
% Leveraging Short-Term Temporal Information to Improve Camera Trap Image Burst Classification
% Leveraging Sequence Information to Improve Camera Trap Image Classification

% Authors at the same institution
%\author{First Author \hspace{2cm} Second Author \\
% Authors at different institutions
\author{
Bhuvan Malladihalli Shashidhara \\
University of Washington\\
{\tt\small msbhuvan@uw.edu}
\and
Darshan Mehta \\
University of Washington\\
{\tt\small darshanm@uw.edu}
\and
Yash Hemant Kale \\
University of Washington\\
{\tt\small yashkale@uw.edu}
\and
Dan Morris \\
Microsoft AI for Earth\\
{\tt\small dan@microsoft.com}
\and
Megan Hazen \\
University of Washington\\
{\tt\small mh75@uw.edu}
}

\maketitle
\ifwacvfinal\thispagestyle{empty}\fi

%%%%%%%%% ABSTRACT
\begin{abstract}
    Motion-triggered wildlife cameras, also known as ``camera traps", are extensively used to observe wildlife in their natural habitat without disturbing the ecosystem. A massive number of such camera traps have been deployed at various ecological conservation areas around the world, and the images from those cameras are typically classified manually by ecologists. Recent efforts to automate this classification with machine learning utilize each of those images independently; however, due to challenging scenes with animals camouflaged in their natural habitat, it is sometimes difficult to identify the presence of animals from a single image. Fortunately, camera traps generally capture ``bursts" of three to five images when an animal passes by the sensor. We hypothesize that leveraging information in this multi-image burst, assuming that the animal moves, makes it much easier to algorithmically detect the presence of animals. In this work, we explore a variety of approaches to extracting information from bursts. We show that concatenating channels containing sequence (temporal) information and the images from the three-image-burst across channels improves the ROC AUC by 20\% on a test set from unseen camera sites when compared to an equivalent model that learns from each image independently.
\end{abstract}

%%%%%%%%% BODY TEXT
\section{Motivation}
Camera trapping is a method to capture wild animals on film without disturbing the ecosystem, and has been used in ecological research for decades \cite{glover2019camera}. A camera trap is a camera equipped with a motion sensor ( typically a passive infrared sensor) and an infrared flash that allows night-time photography with minimal disruption to animal behavior. Most camera traps are configured to capture a ``burst'' of three to five images each time an animal passes by the camera. Images are frequently empty due to false triggers, slow camera wake-up, or animals moving out of frame between images; the largest publicly-available camera trap dataset, for example, contains approximately 75\% empty images \cite{snapshotserengeti}.

Currently, a massive number of camera traps have been deployed at various ecological conservation areas around the world; some of these cameras have been collecting data for decades. Wildlife researchers are interested in performing ecological studies like population estimation of different animal species across different seasons of a particular ecological conservation area, which requires annotating camera trap images, typically to indicate the number and type of all animals present in each image. Due to the vast number of images generated by camera trap surveys, image review is often very time-consuming for researchers, requiring conservation organizations to invest time and resources in data annotation that could be allocated to conservation planning. Consequently, automating the process of labeling camera trap images has become an active area of computer vision research.

The substantial majority of this research, however, performs classification on a single image at a time. However, due to challenging scenes with animals camouflaged in their natural habitat, and due to poor illumination at night, it is often difficult to detect and classify animals from merely a single image. This is particularly pronounced for small animals in night-time images, where motion may be the only visual evidence of an animal's presence. For example, rabbits frequently appear as little more than two bright dots (the animal's eyes reflecting the camera's flash) in a single image, which may be indistinguishable from noise, but over a short burst (sequence) of images, we can see these bright dots move in a pattern which enables us to identify the source as a rabbit.

We hypothesize that utilizing the temporal information present in a burst of images, assuming that the animal moves, it becomes much easier for machines to detect the presence of an animal. Hence, in this research project, we explore and measure the impact of using image sequences on improving the camera trap image classification.

% ----------------------------------------------------------------------------

This paper initially provides a brief review of prior literature, then establishes the problem statement along with the dataset description. This is followed by a detailed description of the proposed methodology, and the discussion of results.

% ----------------------------------------------------------------------------

\section{Literature Review} \label{section:lit-review}

\subsection{Classification of Single Camera Trap Images}

The majority of work on automating camera trap image classification has leveraged only a single image at a time.  For example, \cite{norouzzadeh2018automatically} use three CNNs to annotate camera trap images: one each for presence detection, species classification, and counting. Willi et al. \cite{cnn_camera_trap} similarly use two CNNs: one each for presence detection and species classification. Both of these systems achieve high accuracy, but use the same camera locations in training and test, which later work showed does not allow for generalization of results to new cameras \cite{beery2018recognition}. A complete survey of single-image classification approaches is beyond the scope of this section; we refer the reader to \cite{tabak2019machine} for a more in-depth review.

\subsection{Classification of Camera Trap Bursts}

Though most prior work in this area has focused on single-image classification, several projects have explored the use of temporal information to improve camera trap burst classification. 

Wei et al. \cite{wei2020zilong} use  pixel-level differences between frames to measure color change and the Sobel algorithm to identify edges, then threshold changes in color and contour to determine emptiness.  While this approach is extremely fast, it performed poorly in scenarios where empty images contained swinging vegetation, since these are usually associated with color and edge changes. 

Yousif et al. \cite{animal-scanner} use both classical computer vision algorithms and deep learning to detect and classify moving objects. Specifically, they use differences of histograms of oriented gradients (HOG) and a threshold on intensity entropy to detect candidate moving regions, then use a deep network to classify those moving regions into animal, human, and background.

Giraldo-Zuluaga et al. \cite{pca_camera_trap} propose a multi-layer robust principal component analysis (multi-layer RPCA) approach for background subtraction in camera trap images. Their method computes sparse and low-rank images from a weighted sum of descriptors, using color and texture features. The segmentation algorithm is composed of histogram equalization or Gaussian filtering followed by morphological filters with active contour. They optimize the parameters of their multi-layer RPCA using an exhaustive search. 

\subsection{Other Approaches to Image Sequence Classification}

To exploit the sequence information present in image sequences, Donahue et al. \cite{lstmcnn} train a CNN to extract features from individual images, then use those features as the input to an LSTM that generates sequence-level classifications.  Though not specifically related to camera traps, this work inspired our exploration of recurrent networks for camera trap applications.

% ----------------------------------------------------------------------------

\section{Problem Statement} \label{section:problem_statement}
In this project, we focus on a binary classification task to predict whether a burst (sequence) of three images contains an animal or not. In order to determine whether the sequence information helps, we consider the following two categories of models:
\begin{itemize}
\item \textbf{Baseline Models}: Single-image-based models that learn from a single-image input (single image from the burst of images), and does not utilize any sequence information.
\item \textbf{Sequence Models}: Models that utilize sequence information from the burst of images in some way during the training and inference processes.
\end{itemize}

Our main aim is to determine whether the addition of sequence information improves the image classification. In other words, we aim to test whether the Sequence Models perform better than the Baseline Models on the test set. We use the area under the ROC curve as our primary evaluation metric.

Further, we consider two scenarios for evaluation based on the way we split the data for training and testing: \textit{uniform split} and \textit{site-based split}. Firstly, for the \textit{uniform split} scenario, we assume that data from every camera site is available at training time, and we split the data uniformly across all image bursts to form train, validation, and test sets. This is a reasonable assumption for camera trap surveys that maintain long-term networks of stably-positioned cameras, such as the Snapshot Serengeti project \cite{snapshotserengeti}.

However, this assumption is not reasonable for most camera trap surveys, in which individual camera locations are transient; making each location available at training time would require annotation images and re-training a model every time a new site is added.   Consequently, we consider a second scenario of \textit{site-based split}, where we split the data based on camera sites, to form train, validation, and test sets containing image sequences from mutually exclusive camera sites.

\subsection{Dataset Description} \label{section:dataset}
\begin{figure*}[t]
\begin{center}
\resizebox{\textwidth}{!}{
    \begin{tabular}{lll|lll}
    \includegraphics[width=\linewidth]{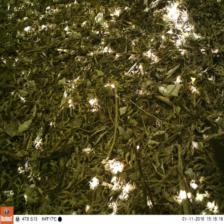} &
    \includegraphics[width=\linewidth]{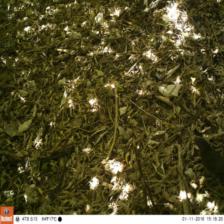} &
    \includegraphics[width=\linewidth]{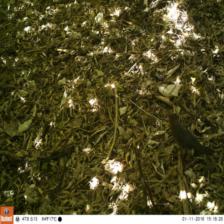}

    & \includegraphics[width=\linewidth]{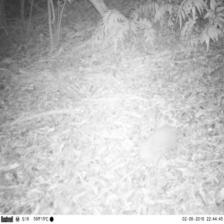} &
    \includegraphics[width=\linewidth]{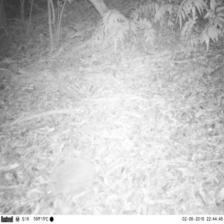} &
    \includegraphics[width=\linewidth]{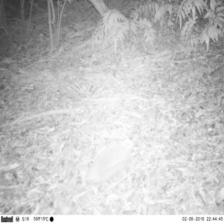}

    \\ \hline

    \includegraphics[width=\linewidth]{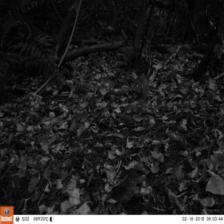} &
    \includegraphics[width=\linewidth]{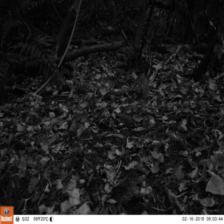} &
    \includegraphics[width=\linewidth]{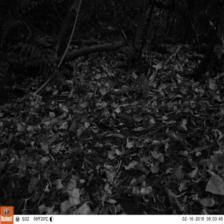}

    & \includegraphics[width=\linewidth]{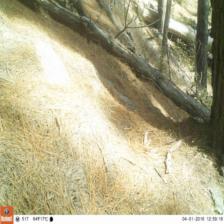} &
    \includegraphics[width=\linewidth]{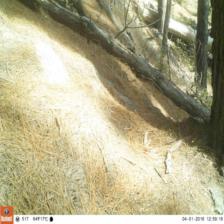} &
    \includegraphics[width=\linewidth]{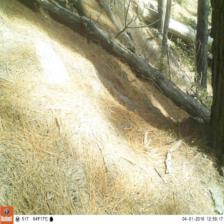}
    \end{tabular}
}
\end{center}
  \caption{Four bursts from Wellington Camera Traps dataset. The bursts contain a bird found in daylight (top-left), a bird in low light (bottom-left), a hedgehog at night (upper-right); each of these animals is hard to spot by looking at a single image, but easier to spot when seen as a burst of images as there is a slight directional animal movement. The sequence on the bottom-right is empty (does not contain any animal), but contains slight movement of leaves due to wind.} \label{fig:sequence_example}
\end{figure*}

The Wellington Camera Traps dataset \cite{wellington} consists of 90150 image bursts collected from camera traps across 187 sites in Wellington, New Zealand; this dataset is publicly available via the Labeled Image Library of Alexandria repository (LILA)\footnote{\url{http://lila.science}}. Each burst consists of three images, leading to a total of 270450 images in the dataset. The data contains bursts that are captured in daylight and at night. This dataset contains numerous small and/or camouflaged animals, so it is frequently difficult to spot an animal by just seeing a single image, but we can frequently spot the animal by observing the burst of images. A few sample sequences from the dataset are shown in Figure \ref{fig:sequence_example}.

Each burst has been annotated into one of 17 classes: 15 classes corresponding to various animal species, the \textit{empty} class for sequences that contain no animals, and the \textit{unclassifiable} class. However, for this study, we binarize the labels to indicate whether the burst contains an animal or not.

% ----------------------------------------------------------------------------
\section{Methodology} \label{section:methodology}
This section describes our data preprocessing steps and the implementation details of both our Baseline Models and Sequence Models.

\subsection{Data Preprocessing}
The Wellington Camera Traps dataset \cite{wellington} is primarily bursts of three images. However, 607 bursts contained fewer than 3 images, so we discarded these bursts to maintain uniformity. We also excluded the 1170 bursts which were annotated as `Unclassifiable'. We binarized the labels to denote whether an animal is present in the sequence or not. All images were resized to $224\times224$ in order to keep training time reasonable.

We create two sets of train, validation, and test sets for the two scenarios described in Section \ref{section:problem_statement}: \textit{uniform split} and \textit{site-based split}. In both scenarios, we oversampled the empty sequences to balance the class distribution in the training set. We also perform down-sampling (random removal) of sequences with animals in order to balance the validation and test sets. For the \textit{uniform split} scenario, the dataset is divided into train (70\%), validation (15\%), and test (15\%) sets by uniform random sampling of bursts, resulting in 62164 bursts for training, out of which 51426 bursts contained an animal while 10738 bursts were empty. After balancing the class distribution as mentioned above, the validation and test sets contained 4278 and 4600 bursts respectively. For the \textit{site-based split} scenario, we considered 111, 36, and 35 sites for the training, validation, and test sets respectively, by random selection. Hence, the training set contained 53855 bursts, out of which 44888 bursts contained an animal while 8967 bursts were empty. After balancing the class distribution as mentioned above, the validation and test sets contained 5968 and 6452 bursts respectively.

\subsection{Implementation Environment}
All of our models are implemented in Python, using TensorFlow 2.0. Code is available on our repository\footnote{\url{https://github.com/bhuvi3/camera_trap_animal_classification} % Comment out for the blind-review and uncomment the below line.
% (foot-note with link to repository would be added in the final version).
}.

\subsection{Baseline Models}\label{section:baseline_model_section}
As introduced in Section \ref{section:problem_statement}, we consider single-image-based models (which do not utilize any temporal information) as Baseline Models. During training, all images from a burst are considered as independent data points. However, during inference, only the first image of each burst is used to obtain the predicted labels, in order to maximize the chance of an animal being present in the frame, since this image is captured as soon as the sensor on the camera gets triggered.

%We also tried taking majority votes from the predictions of each of the 3 images in the sequence, but did not observe any considerable difference.

We compared several standard CNN architectures:  VGG-16 \cite{vgg16}, InceptionResNet-V2 \cite{inceptionresnetv2}, and four variations of ResNet \cite{resnet}, specifically  ResNet-50 \cite{resnet50}, ResNet-101 \cite{resnet101}, ResNet-152 \cite{resnet152}, and ResNet-152-V2 \cite{resnetv2}. We initialized each network with pre-trained weights from ImageNet prior to training to achieve faster convergence.

\subsection{Sequence Models}\label{section:proposed_sequence_models}
\begin{figure}[t]
\begin{center}
  \includegraphics[width=\linewidth]{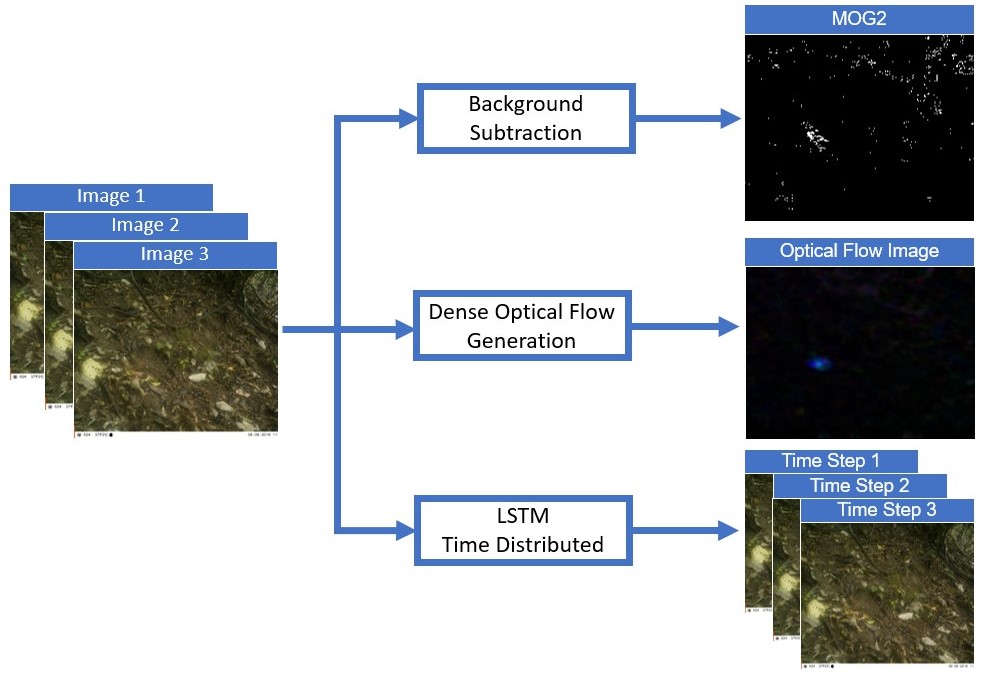}
\end{center}
  \caption{Different approaches for the extraction of sequence information from the three images in a burst). The \textit{Background Subtraction} and \textit{Dense Optical Flow Generation} branches correspond to the Sequence Information Channel Concatenation approach described in Section \ref{section:channel_concat_models}, and the \textit{LSTM Time Distributed} branch corresponds to the Recurrent Neural Architecture approach described in Section \ref{section:rnn}.}
  \label{fig:flowchart}
\end{figure}

As described in Section \ref{section:results}, the ResNet-152 architecture performs well for the Baseline Model. Therefore, ResNet-152 is used as the backbone feature extractor for our Sequence Models. We propose the Sequence Information Channel Concatenation approach in Section \ref{section:channel_concat_models}, and compare it to the Recurrent Neural Architecture approach described in Section \ref{section:rnn}. Figure \ref{fig:flowchart} shows a pictorial overview of the approaches used for Sequence Models.

\subsubsection{Sequence Information Channel Concatenation}\label{section:channel_concat_models}
In this approach, temporal (sequence) information is extracted from the three images of the burst using either the \textit{MOG2} or \textit{Optical Flow} approaches described below. The resulting temporal information is added as additional channels to the first image or to all three images of the burst before they are input to the ResNet-152 model; details on how temporal information is input to the CNN are provided later in this section.

\begin{itemize}
    \item \textbf{MOG2}: This is an OpenCV implementation of a Gaussian-mixture-based background/foreground segmentation algorithm for image sequences \cite{mog2_1,mog2_2}, which produces a grayscale channel capturing the temporal information from the burst of three images. The MOG2 grayscale channel is simply referred as MOG2 in the following sections of the paper.
    \item \textbf{Optical Flow}: We compute dense optical flow between each consecutive image pair in a burst according to Faurneback et al. \cite{opticalflow_gf}, which produces a flow image consisting of RGB channels capturing the temporal information along with directionality. We produce two flow images for the burst of three images: one for optical flow between images 1 and 2 (referred to as ``Optical Flow Image-1-2"), and another for optical flow between images 2 and 3 (referred to as ``Optical Flow Image-2-3"). In some experiments, we use both of these flow images, whereas in other experiments we take the average of these two images to produce a single flow image for the burst of three images (referred to simply as ``Optical Flow Image").
\end{itemize}

\begin{figure*}[t]
\begin{center}
\resizebox{\textwidth}{!}{
    \begin{tabular}{lll|lll}
    \includegraphics[width=\linewidth]{figures/sequence_example/29125_1-1.JPG} &
    \includegraphics[width=\linewidth]{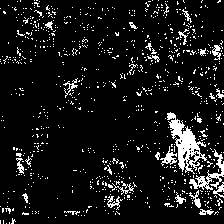} &
    \includegraphics[width=\linewidth]{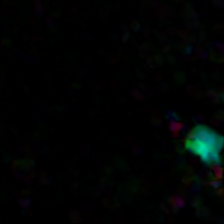}

    & \includegraphics[width=\linewidth]{figures/sequence_example/40156_1-1.JPG} &
    \includegraphics[width=\linewidth]{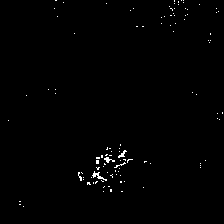} &
    \includegraphics[width=\linewidth]{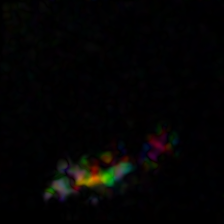}

    \\ \hline

    \includegraphics[width=\linewidth]{figures/sequence_example/17068_1-1.JPG} &
    \includegraphics[width=\linewidth]{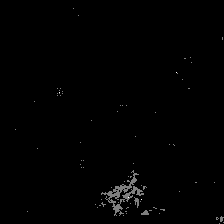} &
    \includegraphics[width=\linewidth]{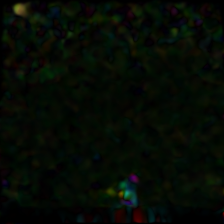}

    & \includegraphics[width=\linewidth]{figures/sequence_example/49305_1-0.JPG} &
    \includegraphics[width=\linewidth]{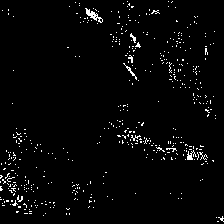} &
    \includegraphics[width=\linewidth]{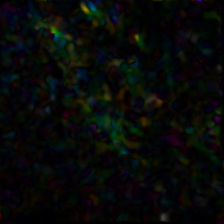}
    \end{tabular}
}
\end{center}
  \caption{Comparison of the MOG2 and Optical Flow methods for extracting temporal (sequence) information. The figure shows the temporal information corresponding to the four bursts shown in Figure \ref{fig:sequence_example}; the first image of the burst is followed by the burst's MOG2 (grayscale channel) and the burst's Optical Flow Image.}
  \label{fig:mog2_opticalflow_comparison}
\end{figure*}

A comparison of the temporal information produced by the MOG2 and Optical Flow methods for a few bursts is shown in Figure \ref{fig:mog2_opticalflow_comparison}.

\begin{figure}[t]
\begin{center}
  \includegraphics[width=\linewidth]{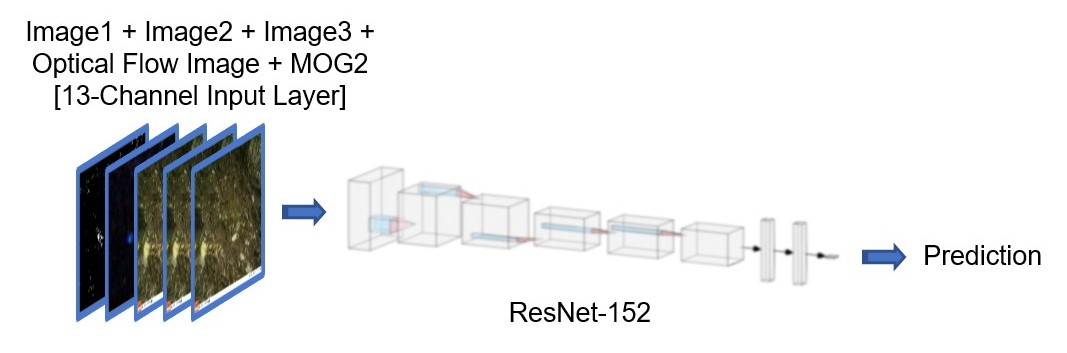}
\end{center}
  \caption{Concatenated input layer along with the backbone CNN architecture (ResNet-152) shown for the Hybrid-13-Channel model. The other Sequence Information Channel Concatenation models described in Section \ref{section:channel_concat_models} use a similar approach.}
  \label{fig:sample_network_diagram}
\end{figure}

We experiment with a variety of channel-wise concatenation possibilities, by modifying the first (input) layer of the model. We initialize weights for the RGB channels corresponding to the images from the burst and Optical Flow Images with the ImageNet pre-trained weights, but use a random-normal initialization for the channels corresponding to MOG2. Based on this approach, we experimented with the following models having variations in the input layer. Image1, Image2, and Image3 refer to the first, second and third images of the burst respectively. In the following model descriptions, the number shown in the parenthesis following the image name represents the number of channels present in the image. Figure \ref{fig:sample_network_diagram} shows this approach for the Hybrid-13-Channel model described below.

\begin{itemize}
    \item \textbf{MOG2-4-Channel}: Image1 (3) stacked with MOG2 (1) across channels to form a four-channel input layer.
    \item \textbf{MOG2-10-Channel}: Image1 (3), Image2 (3), and Image2 (3) stacked across channels along with MOG2 (1) to form a ten-channel input layer.
    \item \textbf{OpticalFlow-6-Channel}: Image1 (3) stacked with Optical Flow Image (3) across channels to form a six-channel input layer.
    \item \textbf{OpticalFlow-15-Channel}: Image1 (3), Image2 (3), and Image2 (3) stacked across channels along with the Optical Flow Image-1-2 (3) and Optical Flow Image-2-3 (3) to form a 15-channel input layer.
    \item \textbf{Hybrid-13-Channel}: Image1 (3), Image2 (3), and Image2 (3) stacked across channels along with Optical Flow Image (3) and MOG2 (1) to form a 13-channel input layer.
    \item \textbf{OpticalFlow-Only-6-Channel)}: Optical Flow Image-1-2 (3), and Optical Flow Image-2-3 (3) stacked across channels to form a six-channel input layer.
    \item \textbf{OpticalFlow-MOG2-Only-7-Channel)}: Optical Flow Image-1-2 (3), Optical Flow Image-2-3 (3), and MOG2 (1) stacked across channels to form a seven-channel input layer.
\end{itemize}

\subsubsection{Recurrent Neural Architecture}\label{section:rnn}
The LSTM \cite{lstm} network architecture is proven to work well in learning sequence information, especially in the natural language processing domain. Therefore, we considered this method of learning the sequence information present in our image bursts.

In our LSTM model, each image in a burst is considered as a time step for the LSTM layer. A TimeDistributed wrapper in TensorFlow \cite{tf_time_distributed} is used over each image in the burst, such that the base CNN architecture is used for each image in the burst. Thus, each image in the burst passes through the base CNN to extract features. The features of all the images in the bursts are then considered as timestamps and passed on to the LSTM layer. For our LSTM model, the backbone architecture was ResNet-50. This model is referred to as ``LSTM" in the rest of the paper.

%----------------------------------

\subsubsection{Training Settings}
For all the models, we apply random augmentations consistently to all components of the concatenated input layer during training. Each of the following described augmentations are applied independently with a 50\% probability.
\begin{itemize}
    \item Horizontal flip: The input is flipped from left to right.
    \item Color transformations: We change the hue, saturation, brightness, and contrast of the image by a random amount within a fixed margin. The margins are hue: [-0.08, 0.08], saturation: [0.6, 1.6], brightness: [-0.05, 0.05], and contrast: [0.7, 1.3]. Note that the color transformations are not applied to MOG2 (grayscale channel) and Optical Flow Image.
    \item Zoom transformation: We crop the image by a random amount chosen between 2\% to 10\% of the input size, then we resize it back to the original input size.
\end{itemize}

All models were trained for ten epochs (as we have a large training set), using the Adam optimizer with a learning rate of $10^{-4}$. Checkpoints were taken at each quarter of an epoch, and inference was run on the validation set to determine the validation ROC AUC at that checkpoint. The models were early-stopped by monitoring validation ROC AUC at each checkpoint with a patience of three checkpoints. The batch size was based on a GPU capacity of 12 GB (Tesla K80), and was set to 32 or 16 based on the model. The best model was saved at each checkpoint when the model resulted in an improved validation ROC AUC. Under this setting, the Baseline Model took about 13 hours to converge, and the Sequence Models took about six hours to converge.

Each of the above configurations is evaluated on the test sets of both the \textit{uniform split} scenario and the \textit{site-based split} scenario. The ROC AUC on the test sets is compared across the Baseline Model and the different Sequence Models, in order to determine whether sequence information improves classification performance.

% ----------------------------------------------------------------------------

\section{Results and Discussion} \label{section:results}
In this section, we describe the selection of the best-performing Baseline Model, which is used to determine the backbone CNN architecture for the Sequence Models described in Section \ref{section:proposed_sequence_models}. We then compare the performance of different Sequence Models to the Baseline Model.

\begin{table}
\begin{center}
\begin{tabular}{|l|c|}
\hline
\multicolumn{1}{|c|}{\textbf{Model Architecture}} & \textbf{\begin{tabular}[c]{@{}c@{}}Test ROC AUC\\ Uniform Split\end{tabular}} \\ \hline
VGG-16             & 0.596 \\ \hline
ResNet-50          & 0.852 \\ \hline
ResNet-101         & 0.842 \\ \hline
ResNet-152         & 0.857 \\ \hline
ResNet-152-V2      & 0.843 \\ \hline
InceptionResNet-V2 & 0.839 \\ \hline
\end{tabular}
\end{center}
\caption{Performance comparison of CNN architectures for the Baseline Model (single-image-based model) on the test set of the \textit{uniform split} scenario. We observe that ResNet models perform well, with ResNet-152 providing the best performance. Therefore, ResNet-152 is considered as the backbone architecture for Sequence Models.}
\label{tab:baselinemodels_comparison}
\end{table}

\begin{table*}
\begin{center}
\begin{tabular}{|l|l|c|c|}
\hline
\multicolumn{1}{|c|}{\textbf{Model Name}} &
  \multicolumn{1}{c|}{\textbf{Model Description}} &
  \textbf{\begin{tabular}[c]{@{}c@{}}Test ROC AUC\\ Uniform Split\end{tabular}} &
  \textbf{\begin{tabular}[c]{@{}c@{}}Test ROC AUC\\ Site-based Split\end{tabular}} \\ \hline
Baseline               & Single-image-based ResNet-152                                                                                                & 0.86 & 0.72 \\ \hline
LSTM                   & LSTM cells over baseline feature maps                                                                                        & 0.88 & 0.76 \\ \hline
MOG2-4-Channel         & Image1 + MOG2                                                                                                           & 0.93 & 0.90 \\ \hline
MOG2-10-Channel        & Image1 + Image2 + Image3 + MOG2                                                                                         & 0.95 & 0.92 \\ \hline
OpticalFlow-6-Channel  & Image1 + Optical Flow Image                                                                                               & 0.95 & 0.92 \\ \hline
OpticalFlow-15-Channel & \begin{tabular}[c]{@{}l@{}}Image1 + Image2 + Image3 + \\ Optical Flow Image-1-2 +\\ Optical Flow Image-2-3\end{tabular} & 0.95 & 0.91 \\ \hline
Hybrid-13-Channel      & \begin{tabular}[c]{@{}l@{}}Image1 + Image2 + Image3 + \\ Optical Flow Image+\\ MOG2 \end{tabular}                   & 0.96 & 0.92 \\ \hline
\begin{tabular}[c]{@{}l@{}}OpticalFlow-Only\\ (6-Channel)\end{tabular} &
  \begin{tabular}[c]{@{}l@{}}Optical Flow Image-1-2 +\\ Optical Flow Image-2-3\end{tabular} &
  0.90 &
  0.92 \\ \hline
\begin{tabular}[c]{@{}l@{}}OpticalFlow-MOG2-only\\ (7-Channel)\end{tabular} &
  \begin{tabular}[c]{@{}l@{}}Optical Flow Image-1-2 +\\ Optical Flow Image-2-3 +\\ MOG2\end{tabular} &
  0.91 &
  0.93 \\ \hline
\end{tabular}
\end{center}
\caption{Comparison of all model performances on the test sets for both the \textit{uniform split} scenario and the \textit{site-based split} scenario. We find that Sequence Models perform significantly better than the Baseline Model in both scenarios. We observe that the Hybrid-13-Channel model results in high performance in both scenarios. Also, the OpticalFlow-Only and the OpticalFlow-MOG2-only models (which do not have any original image from the burst), outperform the Baseline Model, which implies that the Optical Flow Image and MOG2 (grayscale channel) contain helpful temporal information.}
\label{tab:models_comparison}
\end{table*}

% Note: The "OpticalFlow-6-Channel" without all-pretrained weights was found to provide 0.96 in one of our experiments. Therefore, it can be considered the best model as it performs best with least number of params. However, this improvement observed seems to be quite small, therefore in this paper we are considering only the version where all-pretrained weights are used.

The comparison of the performance of various Baseline Model architectures on the test set of the \textit{uniform split} scenario is summarized in Table \ref{tab:baselinemodels_comparison}. We observe that the ResNet-152 architecture provides the highest ROC AUC of 0.857. Therefore, we consider ResNet-152 as the backbone CNN architecture for our Sequence Models.

Table \ref{tab:models_comparison} shows the performance of all the Sequence Models and the Baseline Model on the test sets for both the \textit{uniform split} scenario and the \textit{site-based split} scenario. We observe a drop in the performance on most of the models (except the OpticalFlow-Only (6-Channel) and OpticalFlow-MOG2-Only (7-Channel) models) in the \textit{site-based split} scenario as compared to the \textit{uniform split} scenario, which can be expected due to the presence of new scenes in the test set from unseen sites \cite{beery2018recognition}. However, in both scenarios, we observe that the channel-concatenated Sequence Models (described in Section \ref{section:channel_concat_models}) significantly outperform the corresponding Baseline Models, as the Hybrid-13-Channel model outperforms the Baseline Model on the test sets of \textit{uniform split} scenario and \textit{site-based split} scenario by 10\% and 20\%, respectively. Note that the channel-concatenated Sequence Models share identical feature extraction layers with Baseline Models, i.e., the same ResNet-152 backbone, but still, the Sequence Models yield significantly better performance. This implies that the sequence information captured in Optical Flow Image and MOG2 (grayscale channel) are effectively utilized by concatenating across the channels in the input layer, to achieve significantly better classification of three-image bursts.

We do \textit{not} observe a considerable difference in the test ROC AUC (in either scenario) between the MOG2-based Sequence Models (MOG2-4-Channel, MOG2-10-Channel) and Optical-Flow-Image-based Sequence Models (OpticalFlow-6-Channel, OpticalFlow-15-Channel) when concatenated across channels with the first image (or all images) from the burst. Further, we observe that the OpticalFlow-Only (6-Channel) and OpticalFlow-MOG2-Only (7-Channel) models outperform their corresponding Baseline Models even though they do \textit{not} utilize any original image from the burst, and their input layers contain just the Optical Flow Image and MOG2 (grayscale channel). Also, we observe that Optical Flow Image and MOG2 (grayscale channel) complement each other to provide better performance when both are concatenated together across the channels in the input layer, as seen in the Hybrid-13-Channel model and the OpticalFlow-MOG2-Only (7-Channel) model.

The LSTM model provides a small improvement over the Baseline Model in both scenarios, but it is less effective compared to the channel-concatenated Sequence Models. This could imply that the LSTM models are less effective for short sequences, however, they may be more effective for sequences of longer length like the ones that are usually found in the Natural Language Processing domain, and may merit further exploration for camera traps capable of capturing high-frame-rate videos.

Practically, the \textit{site-based split} scenario is more representative of most deployments, as it corresponds to the utilization of models for sites unseen during training. We observed that Sequence Models offered a significant improvement of about 20\% in test ROC AUC compared to Baseline Models in this scenario, indicating that temporal information through channel-concatenated Sequence Models is beneficial for image classification for locations unseen during training. Though the OpticalFlow-MOG2-only model provided the highest test ROC AUC of 0.93 in the \textit{site-based split} scenario, we choose the Hybrid-13-Channel model as it provides nearly equivalent performance with an ROC AUC of 0.92, and provides a better performance than other models in the \textit{uniform split} scenario.
% Quantitatively, Hybrid-13-Channel model has better performance when averaged across the scenarios. However, the average doesn't really mean anything, therefore not explicitly mentioning this point.

% and it may be more robust as it uses an original image from the burst as well. % This sentence was removed because we do not have quantitative proof that including original image is useful.

%Overall, we think that the channel concatenation configuration in which both the Optical Flow Image and MOG2 (grayscale channel) are concatenated with an original image from the burst, like in the Hybrid-13-Channel model, is best-suited for our problem, as it uses different forms of information (original image from the burst along with Optical Flow Image and MOG2), and because it provides high ROC AUC on the test sets of both scenarios, especially in the \textit{uniform split} scenario where it provides the best ROC AUC of 0.96.

% ----------------------------------------------------------------------------

\section{Conclusion and Future Work}
As our primary contribution in this paper, we demonstrate that the sequence information present in three-image-bursts can be exploited with a simple channel-concatenation approach to significantly improve the classification of camera trap bursts. This is consistent with the subjective observation that animals may be hard to spot in individual images, but are more clearly visible when observed in a short burst of images.

Specifically, we observe that sequence information concatenated across channels with the input image provides a significant improvement of about 10\% and 20\% in the test ROC AUC over a baseline single-image model with the same base CNN architecture, in the \textit{uniform split} scenario and the \textit{site-based split} scenario, respectively. The Optical Flow Image and MOG2 (grayscale channel) seem to be almost equally effective when used independently, but they provide better performance when they are used together. The \textbf{Hybrid-13-Channel} model listed in Table \ref{tab:models_comparison} can be considered as a well-suited Sequence Model for this application as it provides high test ROC AUC in both scenarios. 

Future research can expand this work to test whether similar improvement can be observed in a multi-class species classification setting. Future work can also consider determining the optimal length of the sequence (number of images in the burst), by experimenting with sequences of different length, which might be helpful for optimizing camera trap deployments for ML-based processing. Further, this work could be expanded to determine whether concatenating sequence-information channels helps in an object detection setting where models utilize bounding box or segmentation information.

% ----------------------------------------------------------------------------

\section{Acknowledgment}
%TODO: Anonymize by removing affiliations for blind review.
We thank the Microsoft AI for Earth program for supporting this research project by providing us with cloud credits to perform our experimentation. Also, we would like to thank the Master of Science in Data Science program at the University of Washington for supporting us through the capstone project.

%--- Changes required for Blind Review---------
%- L18: comment out wacv final copy.
%- Ensure author section is removed.
%- L181 (section 4.2). Remove the link to our repository in the footnote.
%- Remove the text in the Acknowledgment section.
%- Check for TODO tags.

% ----------------------------------------------------------------------------

{\small
\bibliographystyle{ieee}
\bibliography{egbib}

\begin{thebibliography}{10}\itemsep=-1pt

\bibitem{wellington}
V.~Anton, S.~Hartley, A.~Geldenhuis, and H.~U. Wittmer.
\newblock Monitoring the mammalian fauna of urban areas using remote cameras
  and citizen science.
\newblock {\em Journal of Urban Ecology}, 4(1):juy002, 2018.

\bibitem{beery2018recognition}
S.~Beery, G.~Van~Horn, and P.~Perona.
\newblock Recognition in terra incognita.
\newblock In {\em Proceedings of the European Conference on Computer Vision
  (ECCV)}, pages 456--473, 2018.

\bibitem{lstmcnn}
J.~Donahue, L.~Anne~Hendricks, S.~Guadarrama, M.~Rohrbach, S.~Venugopalan,
  K.~Saenko, and T.~Darrell.
\newblock Long-term recurrent convolutional networks for visual recognition and
  description.
\newblock In {\em Proceedings of the IEEE conference on computer vision and
  pattern recognition}, pages 2625--2634, 2015.

\bibitem{opticalflow_gf}
G.~Farneb{\"a}ck.
\newblock Two-frame motion estimation based on polynomial expansion.
\newblock In {\em Scandinavian conference on Image analysis}, pages 363--370.
  Springer, 2003.

\bibitem{pca_camera_trap}
J.-H. Giraldo-Zuluaga, A.~Salazar, A.~Gomez, and A.~Diaz-Pulido.
\newblock Camera-trap images segmentation using multi-layer robust principal
  component analysis.
\newblock {\em The Visual Computer}, 35(3):335--347, 2019.

\bibitem{glover2019camera}
P.~Glover-Kapfer, C.~A. Soto-Navarro, and O.~R. Wearn.
\newblock Camera-trapping version 3.0: current constraints and future
  priorities for development.
\newblock {\em Remote Sensing in Ecology and Conservation}, 5(3):209--223,
  2019.

\bibitem{resnet}
K.~He, X.~Zhang, S.~Ren, and J.~Sun.
\newblock Deep residual learning for image recognition.
\newblock In {\em The IEEE Conference on Computer Vision and Pattern
  Recognition (CVPR)}, June 2016.

\bibitem{resnetv2}
K.~He, X.~Zhang, S.~Ren, and J.~Sun.
\newblock Identity mappings in deep residual networks.
\newblock In {\em European conference on computer vision}, pages 630--645.
  Springer, 2016.

\bibitem{lstm}
S.~Hochreiter and J.~Schmidhuber.
\newblock Long short-term memory.
\newblock {\em Neural computation}, 9(8):1735--1780, 1997.

\bibitem{norouzzadeh2018automatically}
M.~S. Norouzzadeh, A.~Nguyen, M.~Kosmala, A.~Swanson, M.~S. Palmer, C.~Packer,
  and J.~Clune.
\newblock Automatically identifying, counting, and describing wild animals in
  camera-trap images with deep learning.
\newblock {\em Proceedings of the National Academy of Sciences},
  115(25):E5716--E5725, 2018.

\bibitem{vgg16}
K.~Simonyan and A.~Zisserman.
\newblock Very deep convolutional networks for large-scale image recognition.
\newblock {\em arXiv preprint arXiv:1409.1556}, 2014.

\bibitem{snapshotserengeti}
A.~Swanson, M.~Kosmala, C.~Lintott, R.~Simpson, A.~Smith, and C.~Packer.
\newblock Snapshot serengeti, high-frequency annotated camera trap images of 40
  mammalian species in an african savanna.
\newblock {\em Scientific data}, 2:150026, 2015.

\bibitem{inceptionresnetv2}
C.~Szegedy, S.~Ioffe, V.~Vanhoucke, and A.~A. Alemi.
\newblock Inception-v4, inception-resnet and the impact of residual connections
  on learning.
\newblock In {\em Thirty-first AAAI conference on artificial intelligence},
  2017.

\bibitem{tabak2019machine}
M.~A. Tabak, M.~S. Norouzzadeh, D.~W. Wolfson, S.~J. Sweeney, K.~C.
  VerCauteren, N.~P. Snow, J.~M. Halseth, P.~A. Di~Salvo, J.~S. Lewis, M.~D.
  White, et~al.
\newblock Machine learning to classify animal species in camera trap images:
  applications in ecology.
\newblock {\em Methods in Ecology and Evolution}, 10(4):585--590, 2019.

\bibitem{resnet101}
TensorFlow.
\newblock Tensorflow keras applications - resnet101., 2020.

\bibitem{resnet152}
TensorFlow.
\newblock Tensorflow keras applications - resnet152., 2020.

\bibitem{resnet50}
TensorFlow.
\newblock Tensorflow keras applications - resnet50., 2020.

\bibitem{tf_time_distributed}
TensorFlow.
\newblock Tensorflow keras layers timedistributed., 2020.

\bibitem{wei2020zilong}
W.~Wei, G.~Luo, J.~Ran, and J.~Li.
\newblock Zilong: A tool to identify empty images in camera-trap data.
\newblock {\em Ecological Informatics}, 55:101021, 2020.

\bibitem{cnn_camera_trap}
M.~Willi, R.~T. Pitman, A.~W. Cardoso, C.~Locke, A.~Swanson, A.~Boyer,
  M.~Veldthuis, and L.~Fortson.
\newblock Identifying animal species in camera trap images using deep learning
  and citizen science.
\newblock {\em Methods in Ecology and Evolution}, 10(1):80--91, 2019.

\bibitem{animal-scanner}
H.~Yousif, J.~Yuan, R.~Kays, and Z.~He.
\newblock Animal scanner: Software for classifying humans, animals, and empty
  frames in camera trap images.
\newblock {\em Ecology and Evolution}, 9(4):1578--1589, 2019.

\bibitem{mog2_1}
Z.~Zivkovic.
\newblock Improved adaptive gaussian mixture model for background subtraction.
\newblock In {\em Proceedings of the 17th International Conference on Pattern
  Recognition, 2004. ICPR 2004.}, volume~2, pages 28--31. IEEE, 2004.

\bibitem{mog2_2}
Z.~Zivkovic and F.~Van Der~Heijden.
\newblock Efficient adaptive density estimation per image pixel for the task of
  background subtraction.
\newblock {\em Pattern recognition letters}, 27(7):773--780, 2006.

\end{thebibliography}
}

\end{document}